\documentclass[runningheads]{llncs}

\usepackage[T1]{fontenc}
\usepackage{graphicx} 
\usepackage{amsmath, amssymb}
\usepackage{array}
\usepackage{url}

\usepackage[]{todonotes}
\usepackage{xcolor}
\usepackage[most]{tcolorbox}

\usepackage{hyperref}

\title{Renormalising Generative Models for Active Inference: Foundations, Derivations, and Verification}
\titlerunning{Renormalising Generative Models for Active Inference}

\author{
Karim Zaghw\inst{1} \and
Andrew Pashea\inst{2} \and
Marc Pritsch\inst{3} \and
Wouter Nuijten\inst{4} \and
Karl Friston\inst{5} \and
Lancelot Da Costa\inst{6}
}

\authorrunning{K. Zaghw et al.}

\institute{
University of Tübingen, Tübingen, Germany
\and
Division of the Social Sciences, University of Chicago, Chicago, IL, USA
\and
Heidelberg University, Heidelberg, Germany
\and
Department of Electrical Engineering, Eindhoven University of Technology, Eindhoven, The Netherlands
\and
Wellcome Centre for Human Neuroimaging, UCL Queen Square Institute of Neurology,
University College London, London, UK
\and
Max Planck Institute for Intelligent Systems, Tübingen, Germany
}

\begin{document}

\maketitle

\begin{abstract}
Active inference offers a unified framework for perception, learning, and action, but scaling discrete active-inference models to rich spatial and temporal domains remains difficult. Renormalising generative models (RGMs) address this challenge by composing discrete generative models across spatial and temporal scales, coarse-graining lower-level states and paths into higher-level causes for objects, events, and action. However, fully reproducing and adapting the framework remains difficult: the mathematical exposition is compact, and the reference implementations are deeply integrated within specialized software environments, leaving many algorithmic details implicit. This paper addresses these challenges by providing a self-contained, derivation-oriented account of RGMs together with an open, verified implementation. We explain how the hierarchy is built, how beliefs and actions are updated within it, and how information is passed between levels. Where the published equations and implementation differ in emphasis, we make those choices explicit and explain their modelling consequences. By clarifying the theory and separating it from its original implementation context, this work lowers practical barriers to entry and makes RGMs more transparent, auditable, and reproducible, providing a foundation for future quantitative evaluation and development on machine-learning benchmarks.
 
\end{abstract}

\section{Introduction}
Active inference treats perception, learning, and action as parts of one inferential process. An agent uses a generative model to infer hidden causes, update model parameters, and choose paths whose predicted consequences have low expected free energy. Discrete active-inference models make this especially transparent because likelihoods, transitions, preferences, and conjugate learning can be written with categorical distributions and Dirichlet parameters. Their main difficulty is scale: a flat model can represent rich spatial and temporal dependencies in principle, but its states, transitions, and policies quickly become too numerous to understand or use directly.

Renormalising generative models (RGMs) address this scaling problem by composing the same discrete model across spatial and temporal scales~\cite{friston2025pixels}. Local hidden states explain observations, paths select transitions, and higher-level states provide empirical priors over the initial states and paths of lower-level episodes. Lower-level trajectories can therefore be summarised as higher-level causes, while higher-level causes contextualise lower-level inference. The term ``renormalising'' refers to the coarse-graining step that groups local patterns, turns repeated configurations into coarser states, and treats lower-level sequences as higher-level events.

This construction lets perception, structure learning, and planning use one probabilistic language. The same hierarchy that explains images or sequences can predict future outcomes under alternative paths. However, the published presentation is compact, and the reference SPM implementation embeds several clarifying choices in procedural Matlab code~\cite{spmGithub}.

This paper builds on \emph{From pixels to planning}~\cite{friston2025pixels} by giving a guided reconstruction of its RGM framework for readers who want to understand, inspect, or reimplement it. The main text reconstructs the hierarchical generative model, explains inference, planning, parameter learning, and active learning, and describes fast structure learning as the concrete renormalisation operator. The implementation and verification section connects this reconstruction to the reproducible Python codebase and clarifies where the explanatory equations differ from the executable routines. Detailed derivations, extended fixed-point/reference-algorithm comparisons, and the full structure-learning construction are available in the archived supplementary material at \url{https://doi.org/10.5281/zenodo.20533539}.

\section{Renormalising Generative Models}
With the scaling motivation in place, this section turns the RGM idea into a working model. It defines the model and notation, then turns to inference, planning, learning, and fast structure learning.

We use a few terms throughout. A hidden-state factor $f$ has categorical values such as $s^f=i$. A path value $u^f=h$ selects a transition slice $B^f_{\cdot\cdot h}$; it may be a latent dynamical mode or a controllable choice biased by planning.  

\subsection{A Hierarchical Generative Model Across Space and Time}
\label{subsec:rgm_space_time}

Following Friston et al.~\cite{friston2025pixels}, an RGM is easiest to read as a recursive wiring pattern for local discrete temporal models. Each level keeps the same interpretation of hidden states, paths, likelihoods, transitions, and empirical-prior mappings. Lower-level episodes are summarised upward; higher-level beliefs return downward as empirical priors.

\subsubsection{From Local Discrete Models to RGMs}
\label{subsubsec:main_local_discrete_rgms}

The local updates use standard variational-message-passing and Dirichlet--categorical conjugacy for discrete graphical models~\cite{winn_variational_2005,kschischang_factor_2001,parr_neuronal_2019}; Appendix~\ref{app:local_messages_static} records the generic identity used below. We write posterior categorical beliefs with hats, such as $\hat s_\tau$ and $\hat u_\tau$, posterior Dirichlet counts with tildes, such as $\tilde a,\tilde b,\tilde d,\tilde e$, use $\sigma$ for softmax, $\odot$ for the relevant tensor contraction, and $\otimes$ for an outer product.

For a Dirichlet-distributed categorical tensor $X$ with posterior counts $\tilde x$, write $\varphi(\tilde x)_{i\lambda}:=\mathbb E[\log X_{i\lambda}]=\psi(\tilde x_{i\lambda})-\psi(\sum_{i'}\tilde x_{i'\lambda})$. Here $i$ indexes the categorical child, $\lambda$ denotes the remaining parent indices, and $\psi$ is the digamma function. This notation covers likelihoods $A$, transitions $B$, hierarchical state mappings $D$, and hierarchical path mappings $E$.

For a likelihood factor, evidence enters as the message $m_{\uparrow A}=O\odot\varphi(\tilde a)$, while conjugate learning adds expected outcome--state counts, $\tilde a=a+O\otimes \hat s$; a one-factor derivation is given in Appendix~\ref{app:local_messages_static}~\cite{bishop_pattern_2006,koller_probabilistic_2009}. Here $O$ is evidence entering inference, distinct from the posterior predictive outcome $\hat o=\mu(\tilde a)\odot \hat s$ formed after state inference.

\subsubsection{Static Hierarchical Coupling}
\label{subsubsec:main_static_hierarchical_coupling}

A static two-level RGM replaces the ordinary prior over a lower state by a higher-level cause. The likelihood $A$ maps lower states to outcomes, whereas a column $D_{\cdot j}$ is an empirical prior over lower states when $s^{(2)}=j$.

The two state updates, derived in Appendix~\ref{app:local_messages_static}, show the bidirectional role of this link:
\[
q(s^{(1)})
=
\sigma\!\left(m_{\downarrow D}+m_{\uparrow A}\right),
\qquad
q(s^{(2)})
=
\sigma\!\left(\log\pi+m_{\uparrow D}\right),
\]
with
\[
m_{\uparrow A}=O\odot\varphi(\tilde a),
\qquad
m_{\downarrow D}=\varphi(\tilde d)\odot \hat s^{(2)},
\qquad
m_{\uparrow D}=\hat s^{(1)}\odot \varphi(\tilde d).
\]
The lower state combines sensory evidence from $A$ with a descending empirical prior from $D$; the higher state receives an ascending message describing which lower state is supported. The vertical mapping learns by the same expected-count rule, $\tilde d=d+\hat s^{(1)}\otimes \hat s^{(2)}$.

With several modalities or co-parent factors, messages add and contract over the other parent beliefs; the corresponding learning rule adds the observed modality's outer product with its posterior parent beliefs. Appendix~\ref{app:local_messages_static} gives the general formula. The RGM architecture keeps these contractions small by assigning each lower-level child to one vertical parent group.

\subsubsection{Temporal Dynamics and Paths}
\label{subsubsec:main_temporal_dynamics_paths}

Temporal RGMs add trajectories with path-conditioned transitions, as in discrete active-inference models with policies or transition regimes~\cite{dacosta_active_2020}. For a hidden-state factor, a path value $u_\tau=h$ selects a transition slice,
\[
p(s_{\tau+1}=i\mid s_\tau=j,u_\tau=h,B)=B_{ijh}.
\]
For an interior time point, the posterior state combines likelihood, forward, and backward transition messages; the corresponding local-message derivation, first without paths and then with path-conditioned dynamics, is given in Appendices~\ref{app:temporal_messages}:
\[
q(s_\tau)
=
\sigma\!\left(
m_{\uparrow A,\tau}
+
m_{\rightarrow B,\tau}
+
m_{\leftarrow B,\tau}
\right),
\]
where
\[
m_{\uparrow A,\tau}=O_\tau\odot\varphi(\tilde a),
\qquad
m_{\rightarrow B,\tau}=\varphi(\tilde b)\odot \hat s_{\tau-1}\odot \hat u_{\tau-1},
\qquad
m_{\leftarrow B,\tau}=\varphi(\tilde b^\top)\odot \hat s_{\tau+1}\odot \hat u_\tau .
\]
Here $\tilde b^\top$ reads the same transition tensor with current-state and next-state indices exchanged. At boundaries, the missing temporal neighbor is absent; in a hierarchy, a descending $D$ message can supply the initial prior. Transition learning adds path-indexed state-change counts, $\tilde b=b+\sum_{\tau=1}^{T}\hat s_\tau\otimes \hat s_{\tau-1}\otimes \hat u_{\tau-1}$.

The mapping $E$ is the path analogue of $D$: $p(u_0=h\mid s^{\mathrm{parent}}=j,E)=E_{hj}$.
The initial path belief combines this descending message with any local path prior:
\[
\hat u_0
=
\sigma\!\left(m_{\downarrow E}+m_{\mathrm{prior}}\right),
\qquad
m_{\downarrow E}
=
\varphi(\tilde e)\odot \hat s^{\mathrm{parent}} .
\]
Its conjugate update is $\tilde e=e+\hat u_0\otimes \hat s^{\mathrm{parent}}$. Later path beliefs are inferred from transition evidence plus a prior term:
\[
\hat u_{\tau-1}
=
\sigma\!\left(m_{\uparrow B,\tau-1}+m_{\mathrm{prior}}\right),
\qquad
m_{\uparrow B,\tau-1}
=
\hat s_\tau\odot\varphi(\tilde b)\odot \hat s_{\tau-1}.
\]
The message $m_{\uparrow B,\tau-1}$ asks which transition slice best explains the inferred change: $s_\tau$ says where the system is, and $u_\tau$ says which local dynamics carry it. Appendices~\ref{app:local_messages_static} and~\ref{app:temporal_hierarchy} give the corresponding variational derivations.

\subsubsection{The Full Space-Time RGM Partition}
\label{subsubsec:main_space_time_partition}

The equations above describe a general message-passing template; an RGM uses a specific partition of it \cite{friston2025pixels}. Each level is a discrete temporal model with hidden states, paths, likelihoods, transitions, and hierarchical empirical priors. A higher state generates a lower group by predicting that group's initial states through $D$ and initial paths through $E$; the lower group then unfolds under its own $B$ tensors and emits outcomes through $A$.

Vertical groups do not overlap: each lower-level initial state or path receives its empirical prior from one parent group. This keeps $D$ and $E$ low-dimensional, replacing high-order co-parent tensors with matrix-like links between a parent group and its children. Cross-group dependence is moved upward rather than represented by dense lateral couplings at every scale, and temporal coarse graining lets deeper levels encode sequences of sequences rather than only instantaneous states.

\subsubsection{Control Paths and the Link to Planning}
\label{subsubsec:main_control_paths_planning}

Some paths are latent dynamical modes, inferred because they explain observed or predicted state transitions. Other paths are controllable: their prior is shaped by expected free energy rather than only by persistence or a higher-level empirical prior. For a controllable path value $h$,
\[
\pi_G(h)
=
\operatorname{norm}_h\!\left(\pi^u_h e^{-\alpha G_h}\right),
\qquad
P(u)\propto \exp(-G(u)).
\]
where $\operatorname{norm}_h$ normalizes over path values, $\pi^u$ is a baseline path prior, $\alpha$ is a precision, and $G_h$ scores the predicted consequences of selecting path $h$. Lower expected free energy makes a controllable path more probable, while the local path update keeps the same form,
\[
\hat u_{\tau-1}
=
\sigma\!\left(m_{\uparrow B,\tau-1}+\log \pi_G\right).
\]
 
\subsection{Inference, Planning, and Active Learning}
\label{subsec:main_inference_planning_active_learning}

The previous subsection defined the space-time RGM. We now ask how beliefs and actions are updated in that model. Two accounts must be kept distinct. The fixed-point equations in Friston et al.~\cite{friston2025pixels} show the local variational-message-passing structure and its expected-log quantities. By the reference algorithm, we mean the executable SPM routine for online RGM inversion, planning, replay, and parent-child recursion. It predicts the next state, corrects that prediction with the current observation, evaluates future paths, and passes posterior summaries between levels. The two accounts use the same model ingredients, but not the same step-by-step update.

\subsubsection{Fixed-Point Messages and the Reference Filter}
\label{subsubsec:main_reference_filter}

In the fixed-point account, an interior state belief is the softmax over observational, forward-transition, and backward-transition expected-log messages.

The reference implementation uses the same model ingredients in a filtering form. For state factor $f$, it first predicts the current state from the previous state and path beliefs:
\[
\bar s_{\tau,i}^f
=
\sum_{j,h}
\mu(\tilde b^f)_{ijh}\,
\hat s_{\tau-1,j}^f\,
\hat u_{\tau-1,h}^f .
\]
This is a posterior-predictive prior: it uses the posterior mean transition tensor $\mu(\tilde b^f)$ rather than the expected-log tensor $\varphi(\tilde b^f)$. The current observation then corrects this prior. Writing $\mathrm{pa}(g)$ for the hidden-state factors that generate modality $g$, define the joint likelihood
\[
\mathcal L_\tau(s_\tau)
:=
\prod_g
P(O_\tau^g\mid s_{\tau,\mathrm{pa}(g)},\mu(\tilde a^g))
\]
which selects the relevant posterior-mean likelihood entries for one-hot evidence and the corresponding categorical product for soft evidence. The filtering posterior is then
\[
q_\tau(s_\tau)
\propto
\mathcal L_\tau(s_\tau)
\prod_f \bar s_{\tau,s_\tau^f}^f,
\qquad
\hat s_{\tau,i}^f
=
\sum_{s_\tau^{-f}}
q_\tau(s_\tau^f=i,s_\tau^{-f}).
\]
The prediction is factorwise, but the correction can still use joint evidence, because an outcome modality may depend on several hidden factors through its likelihood tensor.

The main contrast is not only that the online filter lacks a future-state message. The fixed-point update averages log probabilities, with terms such as $\mathbb E[\log A]$ and state--path averages of $\mathbb E[\log B]$. The reference filter instead takes logs of posterior-predictive probabilities, such as $\log\mathbb E[A]$ and the log of the corresponding average of $\mathbb E[B]$. Appendix~\ref{app:reference_inference_planning} expands this distinction and its modelling consequences.

\subsubsection{Planning as Prospective Path Evaluation}
\label{subsubsec:main_prospective_path_evaluation}

After the current state has been corrected, planning evaluates candidate future paths. This is the path-selection use of expected free energy in Friston et al.~\cite{friston2025pixels}, implemented in posterior-predictive form in the reference algorithm~\cite{spmGithub}. A policy row $V_k$ specifies one path value $V_{k,f}$ for each controllable factor $f$, so the predicted next state is
\[
\hat s_{\tau+1,i}^{f,k}
=
\sum_j
\mu(\tilde b^f)_{ij,V_{k,f}}\,
\hat s_{\tau,j}^f .
\]
The predicted outcome distribution for modality $g$ is obtained by pushing the predicted parent-state distribution through the likelihood:
\[
\hat o_{\tau+1,o}^{g,k}
=
\sum_{s_{\mathrm{pa}(g)}}
\mu(\tilde a^g)_{o,s_{\mathrm{pa}(g)}}
\prod_{f\in \mathrm{pa}(g)}
\hat s_{\tau+1,s^f}^{f,k}.
\]
The shorthand $Q_k(s_{\mathrm{pa}(g)})$ denotes the joint predicted distribution over the parent states of modality $g$ under policy $k$.

The score is expected free energy~\cite{friston_active_2015,parr_generalised_2019,dacosta_active_2020}. A useful one-step posterior-predictive form is
\[
G(k)
\approx
\underbrace{\sum_g
D_{\mathrm{KL}}\!\left(
\hat o_{\tau+1}^{g,k}
\,
\middle\Vert
\,
p_{\mathrm{pref}}^g
\right)}_{\text{Risk}}
+
\underbrace{\sum_g
\sum_{s_{\mathrm{pa}(g)}}
Q_k(s_{\mathrm{pa}(g)})
H_{\mathrm{amb}}^g(s_{\mathrm{pa}(g)})}_{\text{Ambiguity}}
-
\underbrace{N(k)}_{\text{Epistemic Value}}.
\]

Here $p_{\mathrm{pref}}^g$ is the preferred outcome distribution, and $H_{\mathrm{amb}}^g$ is the entropy of the outcome distribution predicted from a fixed parent-state configuration. The first term is risk, the second is ambiguity, and $N(k)$ collects epistemic value, such as expected information gain about likelihood or transition parameters. Since epistemic value is beneficial, it lowers expected free energy; if novelty is ignored, set $N(k)=0$. The compact equations in Friston et al.~\cite{friston2025pixels} sometimes write the same ingredients with expected-log quantities, while the posterior-predictive form above matches the reference algorithm's use of posterior means~\cite{spmGithub}.

For a single controllable path this gives $\pi_G(h)=\operatorname{norm}_h(\pi_h^u e^{-\alpha G_h})$, where $\pi^u$ is the baseline path prior and $\alpha$ is a planning precision. With several controllable factors, the posterior over joint policies is
\[
q(k)
=
\operatorname{norm}_k\!\left(\pi_0(k)e^{-\alpha G(k)}\right),
\]
and each factor receives the marginal path prior
\[
\pi_G^f(h)
=
\sum_k q(k)\,\mathbf 1[V_{k,f}=h].
\]
This joint-policy step matters because future outcomes can depend on combinations of hidden factors.

\subsubsection{Retrospective Path Inference}
\label{subsubsec:main_retrospective_path_inference}

Prospective planning asks what a path would cause. Retrospective path inference asks the reverse question: after inferring $\hat s_{\tau-1}$ and $\hat s_\tau$, which path best explains that transition? The fixed-point equations average $\log B$ under the neighbouring state beliefs, while the reference algorithm takes the log after averaging posterior-mean transition probabilities~\cite{friston2025pixels,spmGithub}.

The reference implementation uses the same retrospective idea, but with a posterior-predictive transition likelihood:
\[
LL(h)
=
\log
\left(
\sum_{i,j}
\hat s_{\tau,i}\,
\mu(\tilde b)_{ijh}\,
\hat s_{\tau-1,j}
\right),
\qquad
\hat u_{\tau-1}
=
\sigma\!\left(LL+m_{\mathrm{prior}}\right).
\]
The prior term depends on the path type: an initial lower-level path receives a parent-induced prior, an uncontrolled path can persist from the previous path belief, and a controllable path receives the planning prior $\pi_G^f$. In the reference algorithm, the parent-prior case is usually represented by the posterior-predictive $E_{\mathrm{eff}}^f$ prior described below. Again, the implementation-side evidence term is a log posterior-predictive probability, whereas the fixed-point evidence term is an expected-log message.

\subsubsection{Recursive Inference Across Levels}
\label{subsubsec:main_recursive_inference}

The hierarchy is inverted recursively in the reference algorithm~\cite{spmGithub}: a parent state provides empirical priors for a child segment, and the child returns posterior summaries as evidence for the parent.

In the fixed-point account, downward $D$ and $E$ messages are expected-log messages. The reference implementation instead forms ordinary parent-induced priors using posterior means:
\[
D_{\mathrm{parent}}^f(i)
=
\sum_j
\mu(\tilde d^f)_{ij}
\hat s_{\tau,j}^{(n+1)},
\qquad
E_{\mathrm{parent}}^f(h)
=
\sum_j
\mu(\tilde e^f)_{hj}
\hat s_{\tau,j}^{(n+1)}.
\]
These parent priors are multiplied into the child model's local priors:
\[
D_{\mathrm{eff}}^f(i)
\propto
D_{\mathrm{local}}^f(i)\,D_{\mathrm{parent}}^f(i),
\qquad
E_{\mathrm{eff}}^f(h)
\propto
E_{\mathrm{local}}^f(h)\,E_{\mathrm{parent}}^f(h).
\]
Upward, the child's posterior state and path summaries are treated as soft observations for the parent. The fixed-point parent update would use expected-log $D/E$ messages; the reference implementation scores these summaries under posterior-mean mappings and then takes logs. Appendix~\ref{app:reference_inference_planning} gives the full downward/upward comparison, including optional replay.

\subsubsection{Parameter Learning and Active Learning}
\label{subsubsec:main_parameter_active_learning}

Once states and paths have been inferred, parameter learning uses the same conjugate expected-count form as the local Dirichlet--categorical model~\cite{bishop_pattern_2006,koller_probabilistic_2009}. The RGM-specific sufficient statistics are
\[
\begin{alignedat}{2}
\Delta a^g &= O_\tau^g\otimes_{f\in \mathrm{pa}(g)}\hat s_\tau^f,
&\quad
\Delta b^f &= \hat s_\tau^f\otimes\hat s_{\tau-1}^f\otimes\hat u_{\tau-1}^f,
\\
\Delta d^f &= \hat s_0^f\otimes_{p\in \mathrm{pa}(f)}\hat s^p,
&\quad
\Delta e^f &= \hat u_0^f\otimes_{p\in \mathrm{pa}(f)}\hat s^p .
\end{alignedat}
\]

Active learning treats some candidate updates as decisions. Friston et al.~\cite{friston2025pixels} introduce this for likelihood counts using a binary update gate $r\in\{0,1\}$, with candidates $\tilde a_\tau$ and $\tilde a_\tau+\Delta a_\tau$. Let $a^{(r)}$ denote the selected candidate count tensor. The gate is scored as $q(r)=\sigma(-\beta\,G(a^{(r)}))$, where $\beta$ is the precision of this internal learning decision.

The score $G(a^{(r)})$ temporarily reads the candidate likelihood counts as a joint distribution $P_r(o,s)$ over outcomes and the parent-state configuration $s$:
\[
G(a^{(r)})
=
-
I_r(O;S)
-
\sum_o
P_r(o)\log p_{\mathrm{pref}}(o),
\]
where $I_r(O;S)$ is mutual information under that candidate joint. The first term favors informative likelihoods; the second favors candidates whose induced outcomes agree with preferences, so a valid conjugate increment can still be rejected. Appendix~\ref{app:parameter_active_learning} gives the gate derivation.

Average over the two possible update gates to get the practical rule $\tilde a_{\tau+1}=\tilde a_\tau+q(r=1)\Delta a_\tau$. An optional memory scale $\eta$ can damp this update; large $\eta$ makes the mapping change slowly, while small $\eta$ lets admitted updates have a larger effect.

The reference implementation~\cite{spmGithub} extends the same selective-learning logic to transition counts, although the original paper develops the gate most explicitly for likelihood learning~\cite{friston2025pixels}. The corresponding transition novelty is expected information gain about $B^f$: a policy is epistemically valuable when it is expected to sharpen the transition column for the state and path it will visit. In this way, planning can seek information about both the likelihood mapping $A$ and the transition dynamics $B$.

Finally, not all priors are active-learning targets. Fixed preferred outcomes are not learned from data. When a learnable $\tilde c$ appears, it is an empirical outcome prior, not the agent's fixed preference distribution. Other empirical priors, such as $\tilde c,\tilde d,\tilde e$, are updated by ordinary conjugate increments, optionally with the same damping scale, rather than by an expected-free-energy gate.
 
\subsection{Fast Structure Learning as Renormalisation}
\label{subsec:main_fast_structure_learning}

\subsubsection{The renormalisation problem.}
\label{subsubsec:main_fsl_renormalisation_problem}
The previous subsections described the RGM model and its inference scheme once a hierarchy has already been chosen. Fast structure learning addresses the complementary question: how can such a hierarchy be built from a high-dimensional sequence? A site is one variable location at the current level: at the bottom it may be a pixel, voxel, patch, or feature, while at higher levels it is usually one retained lower-level group summarized by state and path beliefs. Given categorical observations $Y^{(n)}_{i,t,c}$ over hierarchy level $n$, site $i$, time $t$, and channel $c$, the constructive RG step learns a level-$n$ model $M^{(n)}$ and produces coarser data for the next level:
\[
\mathcal R_n:
Y^{(n)}
\longmapsto
\big(M^{(n)},Y^{(n+1)}\big).
\]
At the first level, $Y^{(1)}$ may be pixels, features, or other categorical evidence. At higher levels, it is made from posterior state and path summaries produced below. The fast structure-learning implementation in the SPM code~\cite{spmGithub} makes the RG operator in Friston et al.~\cite{friston2025pixels} concrete: it groups variables, compresses local patterns into states, learns transitions and paths, and passes summaries upward. It is a fast constructive approximation to structural selection, not a literal execution of the active-selection equations; Appendix~\ref{app:fast_structure_learning} gives the constructive details.

\subsubsection{Grouping variables by dependence.}
\label{subsubsec:main_fsl_grouping}
The first step is to decide which lower-level sites should share a hidden cause. If a site has several visible channels, the learner first bundles them into a site-level distribution $Z_{i,t}^{(n)}=Y_{i,t,1}^{(n)}\otimes\cdots\otimes Y_{i,t,m_n}^{(n)}$. Thus, at higher levels, grouping is performed over whole lower-level items: a site carries both a state-belief channel $\hat s$ and a path-belief channel $\hat u$. Pairwise co-occurrences over time define a mutual-information matrix $I$, where $I_{ij}$ is large when sites $i$ and $j$ carry shared temporal structure.

A group can then be described by a binary membership vector $z$, giving the schematic objective
\[
\max_z z^\top I z,
\qquad
z_i\in\{0,1\},
\qquad
\sum_i z_i\le dx_n .
\]
Here $dx_n$ is the maximum group size. The exact discrete problem is combinatorial, so the fast routine uses a spectral relaxation: the principal eigenvector of $I$ gives soft membership scores, and the largest entries become a hard group. This step decides which sites will be compressed together; it does not yet define the hidden states.

\subsubsection{From local patterns to hidden states.}
\label{subsubsec:main_fsl_pattern_states}
Once a group $G_k^{(n)}$ has been chosen, the group is treated as a small local system. The key compression is that recurring local patterns become state values: $s^{(n)}_{k,t}=\ell$ means that the visible pattern of group $k$ at time $t$ belongs to the $\ell$th local pattern class. For example, if a binary pair visits $(0,0),(0,1),(0,1),(1,1)$, the induced state sequence is $(1,2,2,3)$: the hidden state is the index of the repeated local configuration, not an extra hidden feature inserted by hand.

For soft evidence, pattern equality is replaced by similarity between probability patterns. The fast constructor bins the soft patterns that occur in the structure-learning sequence and assigns each time point to one class, so its upward summaries are hard one-hot vectors $\hat s^{(n)}_{k,t}=e_\ell$. The broader RGM formalism can represent soft posterior beliefs, but the fast hierarchy builder passes hard state summaries upward after this pattern-class step.

\subsubsection{Learning local likelihoods and transitions.}
\label{subsubsec:main_fsl_likelihoods_transitions}
After state labels have been assigned, the constructor accumulates state--outcome counts for $A$ and state--transition--path counts for $B$, giving $\tilde a=a+N_A$ and $\tilde b=b+N_B$. The posterior means $\mu(\tilde a)$ and $\mu(\tilde b)$ then define the local likelihoods and path-conditioned transition probabilities. A path value indexes a transition rule: one slice is enough when each state has one observed successor, while branching successors are separated into additional slices. Appendix~\ref{app:fast_structure_learning} gives the counting details.

\subsubsection{Passing summaries upward.}
\label{subsubsec:main_fsl_upward_summaries}
Each retained group sends two summaries to the next level: a state belief and a path belief. For selected lower-level transition times $t_\tau$, usually separated by a temporal stride $dt_n$,
\[
Y^{(n+1)}_{k,\tau,1}
=
\hat s^{(n)}_{k,t_\tau},
\qquad
Y^{(n+1)}_{k,\tau,2}
=
\hat u^{(n)}_{k,t_\tau}.
\]
Here $dt_n$ is the number of lower-level transition steps skipped between adjacent higher-level samples. In the fast constructor, both vectors are typically hard one-hot summaries. The recursion is therefore simple: lower-level inference becomes higher-level data, and the next level models state/path summaries rather than raw observations.
Because $\hat u^{(n)}_{k,t_\tau}$ labels the transition beginning at $t_\tau$, higher levels sample lower-level transition times rather than every lower-level state time.

\subsubsection{Linking levels through empirical priors.}
\label{subsubsec:main_fsl_empirical_priors}
The vertical $D$ and $E$ links arise from reading higher-level likelihoods downward. At level $n+1$, the visible channels are lower-level summaries: channel $1$ predicts state summaries and channel $2$ predicts path summaries. Read downward, these likelihoods become empirical priors, $P(s^{(n,\tau)}_{k,0}=i\mid s^{(n+1)}_{\kappa,\tau}=\ell)=D^{(n)}_{k,\kappa}(i,\ell)$ and $P(u^{(n,\tau)}_{k,0}=h\mid s^{(n+1)}_{\kappa,\tau}=\ell)=E^{(n)}_{k,\kappa}(h,\ell)$.
Here $k$ indexes the lower-level child group, $\kappa$ indexes the higher-level parent group, and $\ell$ is the parent state's value. The notation $0$ is local to the lower-level segment predicted by the higher-level state at time $\tau$, not the first time point of the entire sequence. During structure learning, state and path summaries are sent upward as data; during generation or hierarchical inference, the higher-level state predicts those same quantities downward as empirical priors.

This construction is renormalisation in the operational sense used by RGMs. It replaces many microscopic variables by fewer coarse variables while preserving the local predictive structure needed for inference and planning. Spatial dependence determines groups, repeated configurations define coarse states, transition counts define paths, and state/path summaries become data for the next scale. Groups whose dynamics are constant or uninformative may be discarded. The result is a hierarchy that preserves the generalized MDP form at every level while changing the variables on which that form is expressed.
  
\section{Implementation and Verification}
\label{sec:implementation_and_verification}

To make the account above executable, we translated the relevant RGM routines from SPM's DEM toolbox into Python, with supporting routines for inference, planning, grouping, and recursive-model assembly; the translation is available at \url{https://github.com/atomresearch/RGMs/}. 

\paragraph{Demonstrations.}  
The translation reproduces three demonstrations from the SPM library that together exercise every component of Sections~\ref{subsec:rgm_space_time}--\ref{subsec:main_fast_structure_learning}. \emph{Drone VI} shows active vision and planning in a dynamic three-dimensional scene with a moving object, observed through rays in the drone's field of view. Domain factors and functional likelihood mappings keep the model tractable by computing only what the current view requires. Generated state trajectories are passed to fast structure learning (Section~\ref{subsec:main_fast_structure_learning}), and the learned hierarchy is coupled back to the drone's active-vision model, testing both hierarchy construction and inference in the resulting recursive model at scale. \emph{Atari III} builds a hierarchical model of a simplified Pong-like game from pixel-level observations of random play, with separate streams for pixels, rewards, costs, and actions. It also demonstrates online continued structure learning episodes where further structures are determined in reference to goal states (basins of attraction). The hierarchical model is later compressed while preserving action-relevant structure, making it the broadest end-to-end test of the pipeline.

\begin{figure}[t]
  \centering
  \includegraphics[width=0.8\textwidth,keepaspectratio]{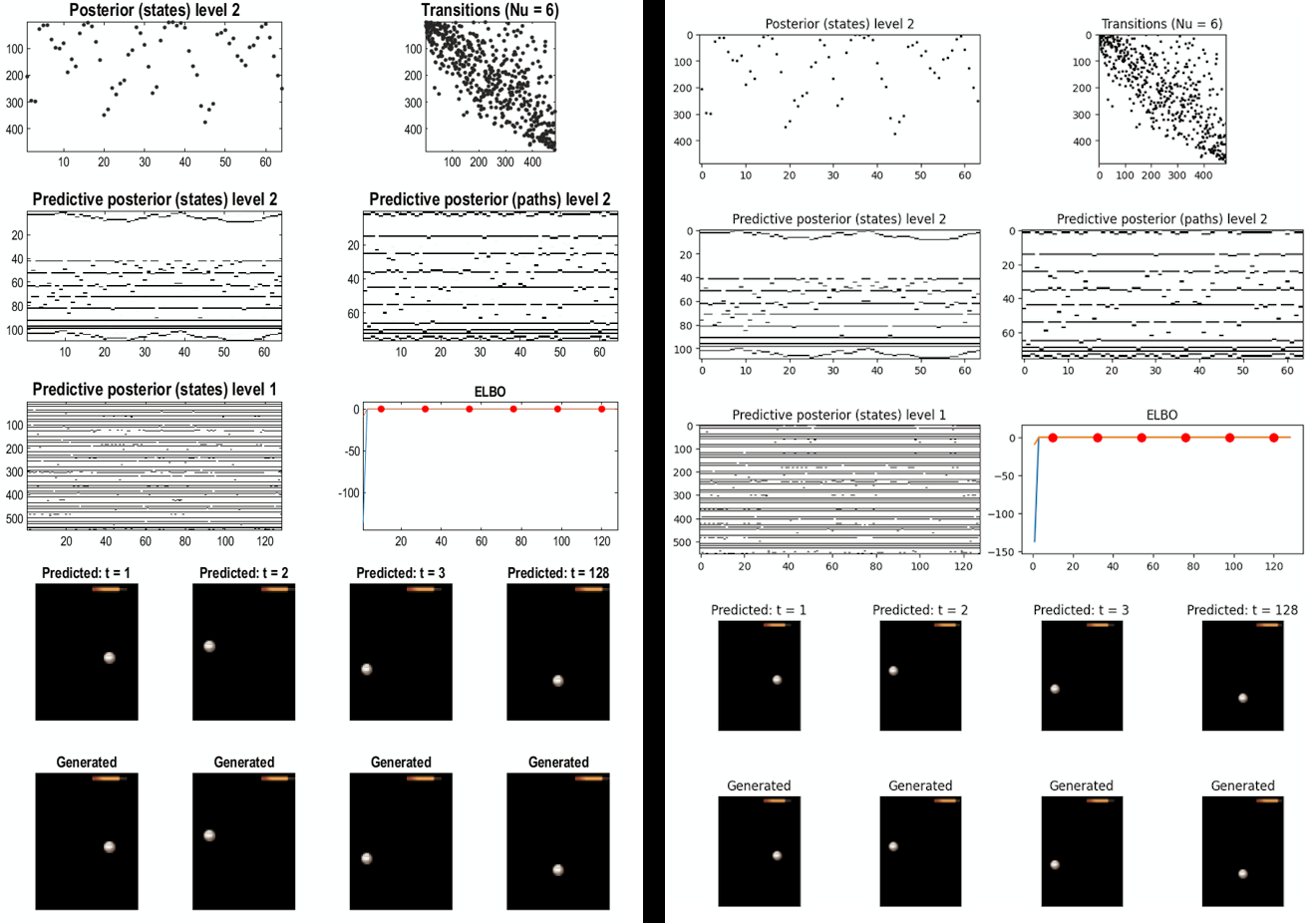}
  \vspace{-4mm}
  \caption{\textit{Left}: MATLAB plot from \textbf{\texttt{DEM\_AtariIII.m}} in the SPM library. \textit{Right}: Python reproduction of the same plot. After fast structure learning and 128 iterations of the reference algorithms, the plot shows the two-level RGM's hierarchical state predictions, retained transitions for a controllable path, ELBO minimization, and predicted versus actual pixel outcomes.}
  \label{fig:matlab-python-atari}
  \vspace{-5mm}
\end{figure}

\paragraph{Oracle testing.}
Verification treated the MATLAB implementation as the executable oracle. The reported verification run passed all tests: the translated Python routines reproduced the MATLAB reference on the checked routines, staged demo probes, and non-visual intermediate outputs. Appendix~\ref{app:verification_methodology} gives the comparison protocol, tolerances, staged-probe design, and precise scope of this claim. The repository should therefore be read as a verified proof-of-concept translation: it establishes fidelity to the SPM reference on the tested surface, rather than optimized runtime performance or a proof that the reference routine is the only possible implementation of the equations. Figure~\ref{fig:matlab-python-atari} shows a representative end-to-end visual check, complementary to the automated oracle tests.

\paragraph{Where the equations and the routines differ.}
This level of detail makes explicit that the reference routines are a tractable online inversion and planning procedure for the model of Section~\ref{subsec:rgm_space_time}, not a literal implementation of every fixed-point equation. Several differences recur: the fixed points use expected logs such as $\varphi(\tilde a)$, whereas the routines take logs of posterior means such as $\log\mathbb E[A]$, weighting parameter uncertainty differently; interior-time fixed points combine likelihood, forward, and backward messages, whereas the routines predict forward and correct on the current observation, yielding a filtering belief; fast structure learning is a constructive approximation to the more general active-selection account of structural change; and the constructor passes hard state and path summaries upward, although the message equations support soft beliefs throughout. The appendix discusses these modelling consequences where they affect the relevant derivations.
 
\section{Conclusion}
RGMs scale discrete active inference by turning lower-level episodes into higher-level causes and returning those causes as empirical priors for lower-level inference and planning. This paper has made that recursive construction explicit through three contributions: a reconstruction of the hierarchical generative model in which every coupling is derived as a local variational message; an explicit separation of these fixed-point equations from the reference algorithm the SPM implementation runs; and an oracle-verified Python translation of the three capstone demonstrations.

Oracle testing demonstrates fidelity to the SPM reference, not independent validation of the RGM framework itself: where the routines approximate the fixed-point equations, the translation reproduces the approximation. As a proof of concept, its coverage is bounded by the three demonstrations.

We see the verified reference as a starting point: natural next steps include benchmarking RGMs against task-specific baselines, optimised reimplementations testable against a known-good reference, relaxing the constructor's hard summaries toward soft beliefs, and a formal treatment of the renormalisation-group reading of fast structure learning. We hope this reconstruction lowers the barrier for such work.
 
\clearpage

\bibliographystyle{splncs04}
\bibliography{references}

\clearpage
\appendix

\section{Appendix}

\subsection{Local Messages and Static Coupling}
\label{app:local_messages_static}

\nameref{subsubsec:main_local_discrete_rgms} and \nameref{subsubsec:main_static_hierarchical_coupling} use local $A$ and $D$ messages as building blocks. This appendix derives those messages without turning the discussion into a full variational-inference tutorial.

\subsubsection{The Local Mean-Field Rule}
\label{app:local_vmp_rule}

The starting point for \nameref{subsubsec:main_local_discrete_rgms} is the generic expected-log identity behind the local RGM messages.

Let the joint density factorise into directed local factors,
\[
p(Z)=\prod_j p(Z_j\mid \mathrm{pa}_j),
\]
and use a mean-field posterior $q(Z)=\prod_i q_i(Z_i)$. Holding all factors except $q_i$ fixed, coordinate ascent on the evidence lower bound gives
\[
\log q_i^*(Z_i)
=
\mathbb E_{q_{-i}}[\log p(Z)]
+\mathrm{const}.
\]
All terms in $\log p(Z)$ that do not involve $Z_i$ are constant with respect to $q_i$. Hence only the factor for $Z_i$ and the factors for its children remain:
\[
\log q_i^*(Z_i)
=
\mathbb E_{q_{-i}}[\log p(Z_i\mid \mathrm{pa}_i)]
+
\sum_{j\in\mathrm{ch}(Z_i)}
\mathbb E_{q_{-i}}[\log p(Z_j\mid Z_i,\mathrm{cp}_j)]
+\mathrm{const}.
\]
This is the only generic variational identity required below. Every RGM update is a softmax over these local expected-log terms, or a conjugate expected-count update for a Dirichlet parameter.

For any categorical tensor $X$ whose columns or parent configurations are Dirichlet-distributed with posterior counts $\tilde x$, define
\[
\varphi(\tilde x)_{i\lambda}
:=
\mathbb E[\log X_{i\lambda}]
=
\psi(\tilde x_{i\lambda})
-
\psi\!\left(\sum_{i'}\tilde x_{i'\lambda}\right).
\]
The index $i$ is the child value and $\lambda$ collects the parent indices. The same notation applies to likelihoods $A$, transitions $B$, state priors $D$, and path priors $E$.

\subsubsection{A Minimal Likelihood Update}
\label{app:minimal_likelihood_update}

The simplest instance is the likelihood message and conjugate $A$ update used in \nameref{subsubsec:main_local_discrete_rgms}.

For one hidden state $s$, one outcome $o$, and likelihood $p(o=i\mid s=j,A)=A_{ij}$, assume $A_{\cdot j}\sim\mathrm{Dir}(a_{\cdot j})$. If $O$ is the evidence vector over outcomes and $\pi$ is the prior over $s$, the state update is
\[
q(s)
=
\sigma\!\left(
\log \pi + O\odot \varphi(\tilde a)
\right).
\]
For a one-hot outcome this selects the row $\mathbb E[\log A_{o^\star j}]$; for soft evidence it averages the same expected-log row under $O$.

The likelihood update follows from conjugacy. Writing $\hat s_j=q(s=j)$,
\[
\log q^*(A)
=
\log p(A)
+
\sum_{i,j} O_i\hat s_j\log A_{ij}
+\mathrm{const},
\]
so each column remains Dirichlet and the posterior counts are
\[
\tilde a=a+O\otimes \hat s,
\qquad
\tilde a_{ij}=a_{ij}+O_i\hat s_j.
\]
The evidence vector $O$ should not be confused with the posterior predictive outcome $\hat o=\mu(\tilde a)\odot \hat s$. The first conditions inference; the second is a prediction formed after inference.

\subsubsection{Static Hierarchical Coupling}
\label{app:static_coupling}

For \nameref{subsubsec:main_static_hierarchical_coupling}, the key point is that a $D$ link acts downward as an empirical prior and upward as evidence for the parent.

A two-level static hierarchy replaces an ordinary lower-level prior with a higher-level cause:
\[
p(o,s^{(1)},s^{(2)},A,D)
=
p(s^{(2)})p(D)p(s^{(1)}\mid s^{(2)},D)p(A)p(o\mid s^{(1)},A).
\]
The likelihood $A$ maps lower states to observations. The hierarchical mapping $D$ maps higher states to lower states, so column $D_{\cdot j}$ is an empirical prior over $s^{(1)}$ when $s^{(2)}=j$.

The two state updates are
\[
q(s^{(1)})
=
\sigma(m_{\downarrow D}+m_{\uparrow A}),
\qquad
q(s^{(2)})
=
\sigma(\log\pi+m_{\uparrow D}),
\]
where
\[
m_{\uparrow A}=O\odot\varphi(\tilde a),
\qquad
m_{\downarrow D}=\varphi(\tilde d)\odot \hat s^{(2)},
\qquad
m_{\uparrow D}=\hat s^{(1)}\odot\varphi(\tilde d).
\]
Thus $D$ has two readings. Downward, it supplies a prior over a lower state. Upward, the inferred lower state supplies evidence about the higher state. Its conjugate update is the same expected-count rule,
\[
\tilde d=d+\hat s^{(1)}\otimes \hat s^{(2)}.
\]

\subsubsection{Multiple Children and Co-Parents}
\label{app:multiple_children_coparents}

The same local rule also covers the multi-parent contractions used in \nameref{subsubsec:main_static_hierarchical_coupling} and \nameref{subsubsec:main_reference_filter}.

If a hidden factor has several child modalities, the child messages add. If modality $g$ depends on several hidden factors, the message to factor $f$ contracts over the observation and over the other parent beliefs:
\[
m_{\uparrow A}^{g,f}
=
O^g\odot\varphi(\tilde a^g)
\odot_{f'\in \mathrm{pa}(g)\setminus f}\hat s^{f'},
\qquad
m_{\uparrow A}^{f}
=
\sum_{g\in\mathrm{ch}(f)}m_{\uparrow A}^{g,f}.
\]
The likelihood-count update is the corresponding outer product,
\[
\tilde a^g
=
a^g+O^g\otimes_{f\in\mathrm{pa}(g)}\hat s^f .
\]
The same algebra applies to hierarchical $D$ links. The main text uses this formula only to justify the small contractions retained by the RGM partition.
 \subsection{Temporal Dynamics and Hierarchical Priors}
\label{app:temporal_hierarchy}

\nameref{subsubsec:main_temporal_dynamics_paths} and \nameref{subsubsec:main_recursive_inference} state the temporal, path-conditioned, and hierarchical-prior messages compactly. This appendix derives those messages and fixes the local time convention used for child segments.

\subsubsection{Temporal Messages Without Paths}
\label{app:temporal_messages}

Begin with the forward and backward transition messages used in \nameref{subsubsec:main_temporal_dynamics_paths}, before paths are introduced.

For one temporal factor without paths,
\[
p(s_{0:T},o_{0:T},A,B)
=
p(A)p(B)p(s_0)
\prod_{\tau=0}^{T}p(o_\tau\mid s_\tau,A)
\prod_{\tau=1}^{T}p(s_\tau\mid s_{\tau-1},B).
\]
At an interior time $\tau$, the local VMP rule keeps exactly three terms: the current likelihood, the transition into $s_\tau$, and the transition out of $s_\tau$. Hence
\[
q(s_\tau)
=
\sigma\!\left(
m_{\uparrow A,\tau}
+
m_{\rightarrow B,\tau}
+
m_{\leftarrow B,\tau}
\right),
\]
with
\[
m_{\uparrow A,\tau}=O_\tau\odot\varphi(\tilde a),
\qquad
m_{\rightarrow B,\tau}=\varphi(\tilde b)\odot\hat s_{\tau-1},
\qquad
m_{\leftarrow B,\tau}=\varphi(\tilde b^\top)\odot\hat s_{\tau+1}.
\]
The notation $\tilde b^\top$ does not introduce a new tensor. It means that the same transition counts are read with current and next-state indices exchanged. At boundaries the missing temporal message is absent; in a hierarchy, the initial prior can instead be supplied by $D$.

Transition learning is the temporal analogue of likelihood learning:
\[
\tilde b
=
b+\sum_{\tau=1}^{T}\hat s_\tau\otimes\hat s_{\tau-1}.
\]
The child index of $B$ is the next state; the parent index is the previous state.

\subsubsection{Path-Conditioned Dynamics}
\label{app:path_conditioned_dynamics}

Adding paths gives the transition messages needed for \nameref{subsubsec:main_temporal_dynamics_paths}, the control interpretation in \nameref{subsubsec:main_control_paths_planning}, and the retrospective path posterior in \nameref{subsubsec:main_retrospective_path_inference}.

Paths turn a single transition matrix into a family of transition rules. If $u_\tau=h$,
\[
p(s_{\tau+1}=i\mid s_\tau=j,u_\tau=h,B)=B_{ijh}.
\]
The state messages therefore average not only over neighbouring state beliefs but also over path beliefs:
\[
m_{\rightarrow B,\tau}
=
\varphi(\tilde b)\odot\hat s_{\tau-1}\odot\hat u_{\tau-1},
\qquad
m_{\leftarrow B,\tau}
=
\varphi(\tilde b^\top)\odot\hat s_{\tau+1}\odot\hat u_\tau .
\]
The transition-count update records the inferred path that explained each state change:
\[
\tilde b
=
b+
\sum_{\tau=1}^{T}
\hat s_\tau\otimes\hat s_{\tau-1}\otimes\hat u_{\tau-1}.
\]

The path posterior asks which transition slice best explains the inferred movement. In the fixed-point account,
\[
\hat u_{\tau-1}
=
\sigma(m_{\uparrow B,\tau-1}+m_{\mathrm{prior}}),
\qquad
m_{\uparrow B,\tau-1}
=
\hat s_\tau\odot\varphi(\tilde b)\odot\hat s_{\tau-1}.
\]
The prior term depends on the path type. It may be a persistence prior, a planning prior for a controllable path, or a parent-induced prior for the initial path of a lower-level segment.

\subsubsection{Hierarchical $D$ and $E$ Priors}
\label{app:hierarchical_de_priors}

The hierarchical part of \nameref{subsubsec:main_temporal_dynamics_paths} and \nameref{subsubsec:main_recursive_inference} depends on a local child-segment convention: $D$ and $E$ are priors over the beginning of that segment.

The mapping $D$ initializes lower-level states, while $E$ initializes lower-level paths:
\[
p(s_0^{(n),f}=i\mid s_\tau^{(n+1)}=j,D^f)=D^f_{ij},
\qquad
p(u_0^{(n),f}=h\mid s_\tau^{(n+1)}=j,E^f)=E^f_{hj}.
\]
In exact fixed-point form, the descending messages are
\[
m_{\downarrow D}^f(i)
=
\sum_j\hat s_{\tau,j}^{(n+1)}\varphi(\tilde d^f)_{ij},
\qquad
m_{\downarrow E}^f(h)
=
\sum_j\hat s_{\tau,j}^{(n+1)}\varphi(\tilde e^f)_{hj}.
\]
The corresponding expected-count updates are
\[
\tilde d^f=d^f+\hat s_0^{(n),f}\otimes\hat s_\tau^{(n+1)},
\qquad
\tilde e^f=e^f+\hat u_0^{(n),f}\otimes\hat s_\tau^{(n+1)}.
\]

The subscript $0$ is local to the lower-level segment generated by the parent state. A higher-level time point $\tau$ selects the beginning of a lower-level segment; within that segment the local initial state is $s_0^{(n)}$ and the local initial path is $u_0^{(n)}$. It is not necessarily the first state or path in the full lower-level sequence.
 \subsection{Fast Structure Learning as the RG Operator}
\label{app:fast_structure_learning}

\nameref{subsec:main_fast_structure_learning} treats the fast constructor as the paper's practical RG operator. The technical construction is given here: grouping variables, compressing local patterns into states, learning path-conditioned dynamics, and sending state/path summaries upward. The source paper also frames structure change in active-selection terms, where alternative structures are compared by evidence and expected-free-energy quantities. Within-family reduction formulas require a shared parameter space, and broader structure comparison requires an explicit candidate family. The fast routine described here is different: it constructs a new coarse model directly from observed dependencies and local pattern classes. The equations below therefore document constructive renormalisation, not a literal execution of the active-selection model-comparison equations.

\subsubsection{Data, Scales, and Site Grouping}
\label{app:fsl_data_grouping}

The construction begins with the level-indexed data, spatial scale, temporal scale, and grouping objective behind \nameref{subsubsec:main_fsl_renormalisation_problem} and \nameref{subsubsec:main_fsl_grouping}.

At level $n$, write the available data as
\[
Y^{(n)}_{i,t,c}\in\Delta^{K_{i,c}^{(n)}-1},
\]
where $i$ indexes a site, $t$ indexes time, and $c$ indexes the visible channel at that site. At the first level, a site may be a pixel, patch, feature, or other categorical evidence source. At higher levels, each retained lower-level group becomes a site with two natural channels: a state-belief channel and a path-belief channel.

Two scales control the step. The spatial scale $dx_n$ bounds the number of sites in a group, and the temporal scale $dt_n$ selects which lower-level transition times are passed upward. The constructive RG map is
\[
\mathcal R_n:Y^{(n)}\mapsto (M^{(n)},Y^{(n+1)}).
\]

Before comparing sites, the channels at each site are bundled:
\[
Z_{i,t}^{(n)}
=
Y_{i,t,1}^{(n)}\otimes\cdots\otimes Y_{i,t,m_n}^{(n)}.
\]
At higher levels this means grouping is performed over whole lower-level summaries, for example over $\hat s_i\otimes\hat u_i$, not over state and path channels as separate objects. Pairwise co-occurrences over time define empirical joint tables and mutual information scores $I_{ij}$. A group can then be selected by the schematic objective
\[
\max_z z^\top I z,
\qquad
z_i\in\{0,1\},
\qquad
\sum_i z_i\le dx_n.
\]
The exact problem is combinatorial. The fast routine uses the principal eigenvector of $I$ as a spectral relaxation, keeps the largest entries up to the group-size bound, removes the selected sites, and repeats. This step chooses which variables share a local hidden cause; it does not yet define the state values of that cause.

\subsubsection{Pattern Classes as Hidden States}
\label{app:fsl_pattern_states}

After grouping, the step described in \nameref{subsubsec:main_fsl_pattern_states} turns observed local patterns into a finite hidden-state alphabet.

After grouping, a group $G_k^{(n)}$ is treated as a small local system. Its visible pattern at time $t$ is the collection
\[
z_{k,t}^{(n)}
=
\{Y_{i,t,c}^{(n)}:i\in G_k^{(n)},\ c=1,\dots,m_n\}.
\]
In the one-hot case, repeated local patterns become state values. If a binary pair visits $(0,0),(0,1),(0,1),(1,1)$, the state sequence is $(1,2,2,3)$. The hidden state is the index of a repeated local configuration, not an extra latent feature introduced independently of the data.

For soft evidence, equality of symbolic patterns is replaced by similarity between probability patterns. The fast constructor compares flattened group patterns, thresholds a distance derived from their normalized inner products, and bins the observed soft patterns into classes. This has two modelling consequences. First, the upward summary produced by the fast constructor is usually hard: $\hat s_{k,t}^{(n)}=e_\ell$. Second, the learned state space contains the pattern classes seen during structure learning. Later inference with the learned model evaluates new evidence under the learned likelihoods; it does not automatically add a new hidden state.

\subsubsection{Likelihoods, Transitions, and Paths}
\label{app:fsl_counts_paths}

\nameref{subsubsec:main_fsl_likelihoods_transitions} rests on two local constructions: count accumulation for $A$ and the interpretation of $B$ slices as empirical transition rules.

Once each time point has a state label, the likelihood counts ask what each visible site and channel tends to look like in each pattern class:
\[
\tilde a_{k,i,c}^{(n)}(y,\ell)
=
a_{k,i,c}^{(n)}(y,\ell)
+
\sum_{t:s_{k,t}^{(n)}=\ell}
Y_{i,t,c}^{(n)}(y).
\]
The state label is hard in the fast constructor, but the accumulated evidence can still be fractional. The learned likelihood column is therefore an average of all soft observations assigned to that pattern class.

The temporal part is learned from the induced state sequence. For group $k$,
\[
B_k^{(n)}(s',s,h)
=
P(s_{k,t+1}^{(n)}=s'\mid s_{k,t}^{(n)}=s,u_{k,t}^{(n)}=h).
\]
In the fast construction, a path indexes an empirical transition rule. If a current state always has the same successor, one path slice is enough. If the same current state branches to different successors, additional path slices separate the alternatives. Thus paths give the learned coarse state space local dynamics.

\subsubsection{How $D$ and $E$ Arise}
\label{app:fsl_de_arise}

The empirical-prior reading in \nameref{subsubsec:main_fsl_empirical_priors} follows from how higher-level likelihoods are defined over lower-level summaries.

At level $n+1$, the visible variables are lower-level summaries. A higher-level likelihood for child site $k$, parent group $\kappa$, and channel $c$ has the form
\[
P(Y_{k,c}^{(n+1)}=y\mid s_\kappa^{(n+1)}=\ell,A_{\kappa,k,c}^{(n+1)})
=
A_{\kappa,k,c}^{(n+1)}(y,\ell).
\]
For the state-summary channel, this likelihood is read downward as a prior over the lower-level initial state:
\[
\tilde d_{k,\kappa}^{(n)}(s,\ell)
\equiv
\tilde a_{\kappa,k,1}^{(n+1)}(s,\ell),
\qquad
P(s_{k,0}^{(n,\tau)}=s\mid s_{\kappa,\tau}^{(n+1)}=\ell)
=
D_{k,\kappa}^{(n)}(s,\ell).
\]
For the path-summary channel, the same reading gives
\[
\tilde e_{k,\kappa}^{(n)}(h,\ell)
\equiv
\tilde a_{\kappa,k,2}^{(n+1)}(h,\ell),
\qquad
P(u_{k,0}^{(n,\tau)}=h\mid s_{\kappa,\tau}^{(n+1)}=\ell)
=
E_{k,\kappa}^{(n)}(h,\ell).
\]
Thus $D$ and $E$ are ordinary likelihood tables at the higher level, but their visible variables are lower-level latent summaries. During structure learning they are learned upward from summaries; during generation or recursive inference they are read downward as empirical priors over the beginning of child segments.
 \subsection{Reference Inference and Planning}
\label{app:reference_inference_planning}

\nameref{subsec:main_inference_planning_active_learning} separates fixed-point messages from the executable reference algorithm. This appendix spells out the main contrasts that the paper body only summarizes: expected-log fixed points, posterior-predictive filtering, and parent-child recursion.

\subsubsection{Filtering Versus Fixed-Point State Messages}
\label{app:reference_filter}

The key comparison for \nameref{subsubsec:main_reference_filter} is the difference between expected-log state messages and posterior-predictive filtering.

In fixed-point form, an interior state receives expected-log likelihood and transition messages, for example
\[
\log q_{\mathrm{VMP}}(s_\tau^f=i)
=
\mathbb E[\log A_{o^\star i}]
+
\sum_{j,h}\hat s_{\tau-1,j}^f\hat u_{\tau-1,h}^f
\mathbb E[\log B^f_{ijh}]
+\cdots .
\]
The omitted term is the backward message from the future state when smoothing is represented directly in the fixed-point graph.

The reference filter first predicts the current state from the previous filtered state and path beliefs:
\[
\bar s_{\tau,i}^f
=
\sum_{j,h}
\mu(\tilde b^f)_{ijh}\,
\hat s_{\tau-1,j}^f\,
\hat u_{\tau-1,h}^f .
\]
It then corrects this predictive prior using the current observation. If $\mathcal L_\tau(s_\tau)$ is the likelihood of all current modalities under a joint state configuration, evaluated with posterior-mean likelihoods, then
\[
q_\tau(s_\tau)
\propto
\mathcal L_\tau(s_\tau)
\prod_f \bar s_{\tau,s_\tau^f}^f,
\qquad
\hat s_{\tau,i}^f
=
\sum_{s_\tau^{-f}}q_\tau(s_\tau^f=i,s_\tau^{-f}).
\]
The prediction is factorwise, but the correction can use joint evidence when an outcome modality depends on several hidden factors.

The mathematical distinction is therefore not just the absence of a future-state message. The fixed-point update averages log probabilities; the filter takes logs of posterior-predictive probabilities:
\[
\mathbb E[\log A_{o^\star i}]
\quad\hbox{versus}\quad
\log\mathbb E[A_{o^\star i}],
\]
\[
\sum_{j,h}\hat s_j\hat u_h\mathbb E[\log B_{ijh}]
\quad\hbox{versus}\quad
\log\sum_{j,h}\hat s_j\hat u_h\mathbb E[B_{ijh}].
\]
They become close when state/path beliefs are sharp and the Dirichlet posteriors are concentrated.

\subsubsection{Retrospective Path Inference}
\label{app:retrospective_path_inference}

Retrospective path inference has the same split: fixed-point path evidence differs from the reference algorithm's posterior-predictive path likelihood.

For a candidate path value $h$, the fixed-point path evidence averages expected log transition probabilities:
\[
m_{\uparrow B}(h)
=
\sum_{i,j}\hat s_{\tau,i}\hat s_{\tau-1,j}
\mathbb E[\log B_{ijh}].
\]
The reference algorithm instead scores the inferred transition under posterior-mean transitions and then takes the log:
\[
LL(h)
=
\log\left(
\sum_{i,j}
\hat s_{\tau,i}\,
\mu(\tilde b)_{ijh}\,
\hat s_{\tau-1,j}
\right),
\qquad
\hat u_{\tau-1}
=
\sigma(LL+m_{\mathrm{prior}}).
\]

\subsubsection{Recursive Parent-Child Inference}
\label{app:recursive_parent_child_inference}

In \nameref{subsubsec:main_recursive_inference}, the same distinction appears inside parent-child recursion, both downward and upward.

The reference algorithm inverts the hierarchy recursively. A parent state supplies empirical priors for a child segment; the child is inverted; the child returns posterior summaries that become evidence for the parent.

Downward, exact VMP would pass expected-log $D/E$ messages. The reference algorithm instead forms posterior-predictive parent priors:
\[
D_{\mathrm{parent}}^f(i)
=
\sum_j\mu(\tilde d^f)_{ij}\hat s_{\tau,j}^{(n+1)},
\qquad
E_{\mathrm{parent}}^f(h)
=
\sum_j\mu(\tilde e^f)_{hj}\hat s_{\tau,j}^{(n+1)}.
\]
These are multiplied into local child priors and normalized:
\[
D_{\mathrm{eff}}^f(i)\propto D_{\mathrm{local}}^f(i)D_{\mathrm{parent}}^f(i),
\qquad
E_{\mathrm{eff}}^f(h)\propto E_{\mathrm{local}}^f(h)E_{\mathrm{parent}}^f(h).
\]
The parent therefore contextualizes the child rather than overwriting it.

Upward, child summaries are treated as soft observations for the parent. A VMP parent update would use terms such as
\[
\sum_i\hat s_{0,i}^{(n),d}\mathbb E[\log D^d_{ij}],
\]
whereas the reference algorithm scores the same summary by a log posterior-predictive likelihood,
\[
\log\left(
\sum_i\hat s_{0,i}^{(n),d}\mu(\tilde d^d)_{ij}
\right).
\]
The same comparison applies to path summaries through $E$. Optional replay or smoothing can revise earlier filtering beliefs after a segment or episode has been observed, but it does not change the central distinction: fixed-point messages use expected logs; the reference recursion uses posterior-predictive quantities.
 \subsection{Parameter Learning and Active Learning}
\label{app:parameter_active_learning}

\nameref{subsubsec:main_parameter_active_learning} gives the learning rule in compressed form. This appendix derives the likelihood update gate and records the implementation caveats that are not derived in the main text.

\subsubsection{The Likelihood Update Gate}
\label{app:likelihood_update_gate}

The likelihood update gate decides whether a candidate count increment should be admitted.

Friston et al. introduce active learning most explicitly for likelihood counts. Let
\[
\Delta a_\tau=O_\tau\otimes_{f\in\mathrm{pa}}\hat s_\tau^f.
\]
The update gate $r\in\{0,1\}$ selects between two candidate futures:
\[
a^{(0)}=\tilde a_\tau,
\qquad
a^{(1)}=\tilde a_\tau+\Delta a_\tau.
\]
This gate is an internal learning decision, not an external action path. Its posterior is
\[
q(r)=\sigma(-\beta G(a^{(r)})),
\]
where $\beta$ is the precision of the learning decision.

To score a candidate likelihood tensor, the count tensor is temporarily read as a joint distribution over outcomes and state configurations:
\[
P_r(o,s)
=
\frac{a^{(r)}_{os}}{\sum_{o',s'}a^{(r)}_{o's'}}.
\]
This global normalization is not the usual column normalization used to form $\mu(\tilde a)$. It is a temporary reinterpretation that makes the mutual-information term well defined. A compact score is
\[
G(a^{(r)})
=
-
I_r(O;S)
-
\sum_o P_r(o)\log p_{\mathrm{pref}}(o).
\]
The mutual-information term favors informative likelihoods. The preference term favors candidates whose outcome marginal agrees with preferences.

Averaging over the two update gates gives the practical update
\[
\tilde a_{\tau+1}
=
\tilde a_\tau+q(r=1)\Delta a_\tau.
\]
With the optional memory scale $\eta$,
\[
\tilde a_{\tau+1}
=
\frac{[\tilde a_\tau+q(r=1)\Delta a_\tau]\eta}
{\eta+q(r=1)}.
\]
Large $\eta$ makes the mapping change slowly; small $\eta$ lets admitted updates have larger effect. If $q(r=0)=1$, the tensor is unchanged.

\subsubsection{Transition Novelty and Other Priors}
\label{app:transition_novelty_other_priors}

Two caveats matter for \nameref{subsubsec:main_prospective_path_evaluation} and \nameref{subsubsec:main_parameter_active_learning}: transition novelty is an implementation extension, and fixed preferences are not empirical priors.

The reference algorithm extends selective learning to transition counts:
\[
\tilde b_{\tau+1}^f
=
\tilde b_\tau^f+q(r=1)\Delta b^f,
\]
again optionally damped by $\eta$. The corresponding epistemic value is expected information gain about $B^f$. A path is exploratory when it is expected to sharpen transition probabilities for the state and path it will visit. This gives planning two epistemic channels: likelihood novelty about $A$ and transition novelty about $B$.

A learnable $\tilde c$ is an empirical outcome prior, not the agent's fixed preference distribution. Fixed preferences $p_{\mathrm{pref}}(o)$ are part of the agent's objective and are not learned from observations. Empirical priors such as $\tilde c$, $\tilde d$, and $\tilde e$ can be updated by ordinary conjugate increments, usually with memory-scale damping, rather than by an expected-free-energy gate.
 \subsection{Verification Methodology}
\label{app:verification_methodology}

Section~\ref{sec:implementation_and_verification} states the verification claim. This entailed running the Python translations of the original SPM scripts with oracle tests, exposing and comparing the output values and structures of each within a stated floating-point tolerance given typical Python-MATLAB minute floating point differences. For functions containing multi-step or iterative operations, as well as their employment in the simulation scripts covered, this equivalence procedure was carried out for intermediate and final outputs alike.

\subsubsection{Oracle Principle}
\label{app:verification_oracle_principle}

For small routines, the comparison is usually just a scalar, vector, or array. For larger routines, the comparison walks through nested MATLAB and Python objects: structs or dictionaries, cells or lists, sparse matrices, function handles, and model fields. The tests compare the kind of object, field names, shapes, empty values, sparse or dense status, logical status, function-handle names, and numeric contents. A mismatch in any required part fails the test. This is important because later RGM routines depend on representation details, not only on final numbers.

Most numeric comparisons use relative tolerance $10^{-7}$ and absolute tolerance $10^{-12}$. Selected recursive demo summaries use relative tolerance $10^{-6}$ and absolute tolerance $10^{-9}$, because they compare large nested structures after many floating-point operations. A few sampling-trace checks are stricter because they compare the probability vector at the sampling site before replaying the MATLAB sample. Discrete quantities, shapes, field names, sparse/logical status, and function-handle names are checked exactly.

\subsubsection{Staged Demo Probes}
\label{app:verification_staged_probes}

For long demonstrations, a probe is a named point in the source code with a short list of variables to expose. The MATLAB probe saves those variables, and the Python probe is run to the corresponding point before comparison. The list is intentionally compact: it records the objects that define the next stage, rather than every temporary variable in the function.

This design makes failures local. If a downstream summary disagrees, the last matching probe narrows the discrepancy to the next construction or inversion step. It also avoids a common trap in stochastic demos: Python is compared under the same recorded random trajectory, using seeds, MATLAB RNG states, or replayed sampling events where needed. Table~\ref{tab:oracle_verification_summary} lists what was checked, how it was compared, and what counted as a match.

\begin{table}[t]
\centering
\footnotesize
\caption{Summary of what the oracle tests checked.}
\label{tab:oracle_verification_summary}
\begin{tabular}{>{\raggedright\arraybackslash}p{0.16\textwidth}>{\raggedright\arraybackslash}p{0.27\textwidth}>{\raggedright\arraybackslash}p{0.25\textwidth}>{\raggedright\arraybackslash}p{0.22\textwidth}}
\hline
Target & What was checked & How it was compared & What counted as a match \\
\hline
Core DEM routines & Helper outputs for indexing, normalization, grouping, pruning, generation, evidence, and model conversion. & MATLAB and Python were run on the same inputs. & Same output types, fields, shapes, sparse/logical status, and numeric values within $10^{-7}$ relative and $10^{-12}$ absolute tolerance. \\
\hline
Reference inference & Filtering, planning, and recursive inversion (\texttt{spm\_MDP\_VB\_XXX}, \texttt{spm\_VBX}). & Final outputs and selected internal summaries were compared. & Same posterior fields and recursive structures; relaxed tolerance for large recursive summaries only where stated. \\
\hline
Fast structure learning & Grouping, count accumulation, learned likelihoods and transitions, hierarchy construction, and RDP conversion (\texttt{spm\_rgm\_group}, \texttt{spm\_faster\_structure\_learning}). & Constructor and conversion checkpoints were compared. & Same learned dimensions, indices, sparse/logical status, and numeric values within tolerance. \\
\hline
Drone VI & Dynamic scenes, training data, learned models, recursive objects, and inference summaries. & Full and reduced staged probes were compared. & Same fields and summaries; full scene up to the degenerate RG-grouping ordering (cf.\ Atari III).\\
\hline
Atari III & De novo fast structure learning under random play, additional structure merging/pruning via Bayesian model reduction (basins of attraction where only generalised goal states with high NESS are retained and sorted), model compression.\cite{friston_gradient-free_2025}   & Inputs and outputs of all core routines compared at all call sites for extensibility across Atari III and other simulations. & Same recorded fields (comprehensive) and numerical equivalence; random stages used the recorded random trajectory. Python-native RG grouping closely approximates sorting (51/58 groups in Atari III), agent behavior maintained.  \\
\hline
\end{tabular}
\end{table}

\subsubsection{What This Establishes}
\label{app:verification_interpretation}

The protocol establishes equivalence to the MATLAB reference on the checked boundaries. It does not choose among mathematically equivalent implementations of the fixed-point equations, and it does not benchmark runtime. Its value is narrower and more direct: it shows that the translated routines used by the demonstrations return the same objects as the SPM routines at the tested points, including representation choices that affect inference, planning, structure learning, and recursive hierarchy construction.
 
\end{document}